\documentclass[runningheads]{llncs}

\usepackage{eccv}

\usepackage{eccvabbrv}

\usepackage{graphicx}
\usepackage{booktabs}

\usepackage[accsupp]{axessibility}  

\usepackage[breaklinks,colorlinks,citecolor=eccvblue]{hyperref}

\usepackage{orcidlink}

\usepackage{makecell}
\usepackage{multicol,multirow}
\usepackage{tcolorbox}

\begin{document}

\title{QUAR-VLA: Vision-Language-Action Model for Quadruped Robots} 

\titlerunning{QUAR-VLA: Vision-Language-Action Model for Quadruped Robots}

\author{Pengxiang Ding$^{12}$ \quad Han Zhao$^{12}$ 
\quad Wenxuan Song$^{1}$
\quad Wenjie Zhang$^{1}$ \quad Min Zhang$^{12}$  \quad Siteng Huang$^{12}$  \\ \quad Ningxi Yang$^{1}$ \quad Donglin Wang$^{1}$\thanks{Corresponding author}}
\institute{$^{1}$Westlake University \quad $^{2}$Zhejiang University \\ 
\email{\{dingpengxiang,zhaohan34,wangdonglin\}@westlake.edu.cn}}

\maketitle

\newcommand{\zhao}[1]{\textcolor[rgb]{0.9, 0.0, 0.0}{\textbf{#1}}}
\newcommand{\px}[1]{\textcolor{blue}{[Pengxiang: #1]}}
\newcommand{\wx}[1]{\textcolor{orange}{[wx: #1]}}
\newcommand{\wj}[1]{\textcolor{purple}{[wj: #1]}}


\begin{abstract}
The important manifestation of robot intelligence is the ability to naturally interact and autonomously make decisions.
Traditional quadruped robot learning typically handles language interaction and visual autonomous perception separately, which, while simplifying system design, also limits the synergy between different information streams. This separation poses challenges in achieving seamless autonomous reasoning, decision-making, and action execution.
To address these limitations, a novel paradigm, named \textbf{V}ision-\textbf{L}anguage-\textbf{A}ction tasks for \textbf{QUA}druped \textbf{R}obots (\textbf{QUAR-VLA}), has been introduced in this paper. This approach tightly integrates visual information and instructions to generate executable actions, effectively merging perception, planning, and decision-making. The central idea is to elevate the overall intelligence of the robot.
Within this framework, a notable challenge lies in aligning fine-grained instructions with visual perception information. This emphasizes the complexity involved in ensuring that the robot accurately interprets and acts upon detailed instructions in harmony with its visual observations. Consequently, we propose \textbf{QUA}druped \textbf{R}obotic \textbf{T}ransformer (\textbf{QUART}), a VLA model to integrate visual information and instructions from diverse modalities as input and generates executable actions for real-world robots and present \textbf{QUA}druped \textbf{R}obot \textbf{D}ataset (\textbf{QUARD}), a large-scale multi-task dataset including perception, navigation and advanced capability like whole-body manipulation tasks for training \textbf{QUART} model. Our extensive evaluation shows that our approach leads to performant robotic policies and enables \textbf{QUART} to obtain a range of generalization capabilities. 

\end{abstract}

\section{Introduction} \label{sec:introduction}

Quadruped robots, characterized by their excellent traversability on complex terrains and agile movements, have garnered significant attention in the field of robotics~\cite{Hutter2016anymal}. Researchers have extensively employed these robots to explore tasks encompassing autonomous navigation and manipulation~\cite{tang2023saytap, vinl, karnan2022scand}.

\begin{figure}[t]
\begin{center}
  \includegraphics[width=1\columnwidth]{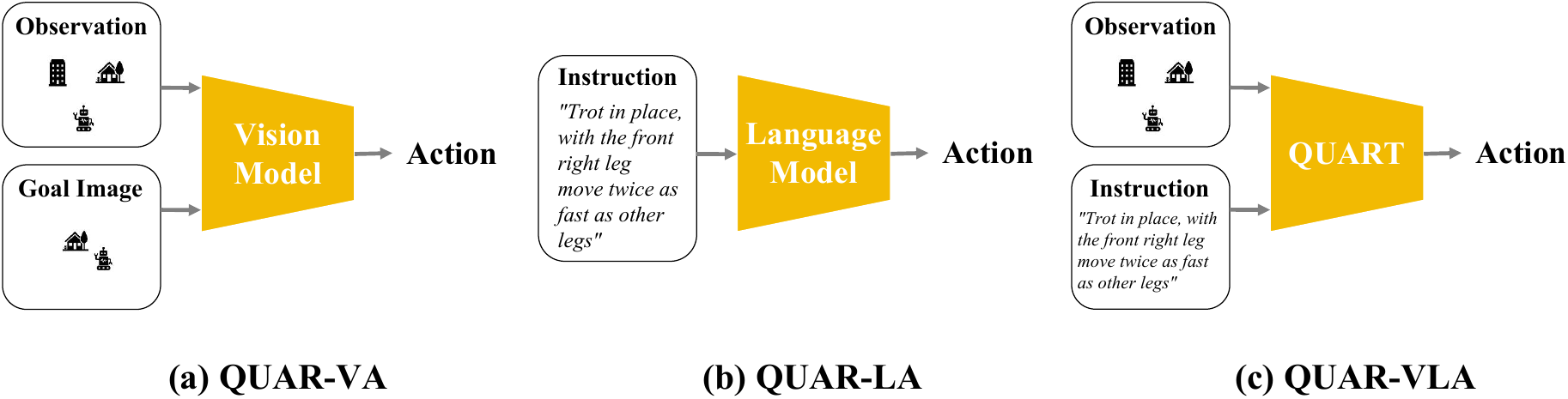} 
\end{center}
\caption{\textbf{Comparison of QUAR-VA, QUAR-LA, and QUAR-VLA.} \textbf{QUAR-VA} solely utilizes coarse-grained vision information, lacking explicit instructions for handling diverse tasks. In contrast, \textbf{QUAR-LA} exclusively relies on language information and lacks of vision information for autonomy. Therefore, \textbf{QUART-VLA} combines both vision information and language instructions as inputs, enabling autonomous problem-solving across a range of tasks, revealing distinct input modalities and task capabilities.}
\vspace{-1.5em}
\label{fig:comparison}
\end{figure}

{Broadly speaking, the quadruped tasks consist of two major specifications:} 
\textbf{V}ision-\textbf{A}ction tasks for \textbf{QUA}druped \textbf{R}obots (\textbf{QUAR-VA}) and \textbf{L}anguage-\textbf{A}ction tasks for \textbf{QUA}druped \textbf{R}obots (\textbf{QUAR-LA}). 
As depicted in Fig.~\ref{fig:comparison}, in \textbf{QUAR-VA} approaches~\cite{shah2023vint}, quadruped robots \{receive\} perception images captured from a first-person perspective and instruction images obtained from a third-person perspective to guide their actions. 
{However, such a task specification often relies on a single (coarse-grained) goal image instruction, making it difficult to apply in many real-world combination tasks, \ie requiring combining multiple sub-instructions.}   
In contrast, employing language as instructions~\cite{tang2023saytap}, {\textbf{QUAR-LA} formulation} allows for executing more fine-grained and diverse tasks, 
{suitable for expressing combinational commands (\ie \textit{``before / then"}),  complex spatial relationships (\ie \textit{``move to the left"}), and many commonsense command priors (\ie \textit{``move fast"}).} 
Nevertheless, \textbf{QUAR-LA} approaches, which lack the integration of visual modality, hinder the robots' ability to perceive the environment, thereby impeding their autonomous navigation capabilities\cite{tang2023saytap}. 
To enable quadruped robots to autonomously navigate and manipulate various tasks, in this paper, we propose a new paradigm: \textbf{V}ision-\textbf{L}anguage-\textbf{A}ction tasks for \textbf{QUA}druped \textbf{R}obots (\textbf{QUAR-VLA}), integrating visual information and instructions from diverse modalities as input and generating executable actions for real-world robots. 

This task primarily encompasses two challenges.
Firstly, there is a lack of large-scale datasets in the research community for quadruped robots performing diverse tasks.
Pretraining models require abundant trajectory data from a variety of tasks. However, collecting substantial amounts of real-world data often necessitates human involvement and significant time investment in manual instruction operations. 
Secondly, different from navigation tasks~\cite{shah2023vint} for general mobile robots and manipulation tasks for fixed-base systems~\cite{stone2023openworld, brohan2023rt1, brohan2023rt2, gu2023rttrajectory}, building a VLA model to solve complex quadruped robot tasks is considerably more challenging due to their agile locomotion behaviors. The action space needs to be properly defined to strike a balance between movement flexibility and computational efficiency. The action generated by the model should not be too simplistic, akin to the planner base velocities output by 2-D navigation modules, nor should they require a high execution frequency like the low-level motion policy that directly controls the joint motors.

To address these two problems, we collect a large-scale multi-task dataset \textbf{QUA}druped \textbf{R}obot \textbf{D}ataset (\textbf{QUARD}). It includes multiple tasks such as perception, navigation, and advanced capabilities like object avoidance. To the best of our knowledge, this is the first quadruped robot dataset that incorporates a significant amount of vision, language instruction, and robot command data. As collecting data on real robots is expensive and inefficient, we primarily rely on data generated in simulation, which exhibits significant differences in visual, sensor, and system dynamics. 
We also introduce \textbf{QUA}druped \textbf{R}obotic \textbf{T}ransformer (\textbf{QUART}), a VLA model for training \textbf{QUARD}. \textbf{QUART} takes images of the robot's first-view camera and the natural language instruction as inputs and generates rich control commands that include controlling the robot's base velocity, posture, and gait parameters.
\textbf{QUART} leverages a pre-trained large-scale visual language model and fine-tunes it on our dataset to enable the generation of executable commands for quadruped robots.
To address the sim-to-real gap caused by the data disparity, we construct a co-training pipeline to effectively distill the knowledge of simulation data for real-scene deployment.
Our extensive evaluation shows that our approach leads to performant robotic policies and enables \textbf{QUART} to obtain a range of generalization capabilities.

\begin{figure}[t]
\begin{center}
      \includegraphics[width=1\columnwidth]{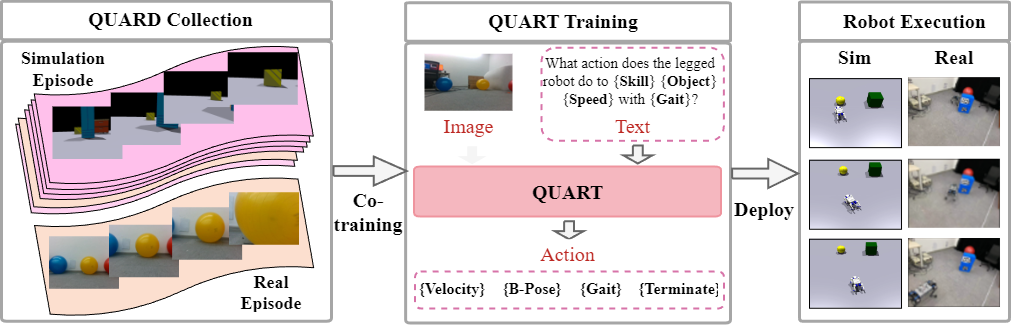} 
\end{center}
\caption{
\textbf{Overview of \textbf{QUAR-VLA}}.
Our tasks encompass a diverse range of perception, navigation, and other advanced capability. The Vision-Language-Action (VLA) model first undergoes training with a huge amount of simulation data (259K episodes) and a small amount of real-world data (3K episodes). In the inference phase, images and texts undergo tokenization, after which \textbf{QUART} generates 12-dimensional action tokens. These tokens are subsequently detokenized into valid robot actions and deployed on a physical quadruped robot. This methodology effectively extends the learned capabilities from a simulated environment to real-world applications.
}
\vspace{-1.em}
\label{fig: teaser}
\end{figure}
Our contributions are as follows:
1) To the best of our knowledge, we first propose a new paradigm: \textbf{QUAR-VLA}, integrating visual and language instructions to executable actions for more autonomous and diverse quadruped robot tasks.
2) We present a large-scale multi-task dataset, \textbf{QUARD}, and a Vision-Language-Action model, \textbf{QUART} to solve the \textbf{QUAR-VLA} tasks.
3) Our extensive evaluation shows that our approach leads to performant robotic policies and enables \textbf{QUART} to obtain a range of generalization capabilities.

\section{Related Work} \label{sec:relatedwork}

\noindent\textbf{Quadruped Robot Learning.} Previous research~\cite{vinl,caluwaerts2023barkour,karnan2022scand,Yang2021LearningVQ,tang2023saytap, pari2021surprising,10160747,bahl2023affordances,lee2019icra,belkhale2023hydra,haldar2022watch} has delved into single-model policies. As to vision-based methods (\textbf{QUAR-VA}), Kareer~\etal~\cite{vinl} utilizes vision perception to predict a privileged terrain map, enabling the robot to perform specific locomotion based on the terrain characteristics. Karnan~\cite{karnan2022scand} leverages ego-motion estimates obtained through vision perception, allowing the robot to understand its relative position and direction to avoid collisions and navigate toward a goal location.
While some works have succeeded in these tasks through vision perception, relying solely on vision limits the capability to handle only a specific type of task.
In response to this limitation, Tang~\etal~\cite{tang2023saytap} have started exploring language-conditioned tasks (\textbf{QUAR-LA}). 
However, the absence of vision perception in their approach results in a lack of autonomy for the robot to interact with its environment. 
Therefore, this paper proposes a new \textbf{QUAR-VLA} paradigm to integrate visual information and instructions to generate executable actions, effectively ensuring that the robot accurately interprets and acts upon detailed instructions in harmony with its visual observations.

\noindent\textbf{Vision-language-action models.}
Visual-language-action models~\cite{shridhar2022cliport,reed2022gato,Nair2022R3MAU,li2023vision,szot2023llarp, bharadhwaj2023roboagent, gdm2024autort}integrates visual information and instructions to generate executable actions, effectively merging perception, planning, and decision-making and elevating the overall intelligence of the robot. 
As a result, vision-language-action models have been a popular area of research in robotics.
In the realm of visual-language-action models, Shridhar~\etal~\cite{shridhar2022cliport} have made significant strides by leveraging CLIP to encode text inputs, thereby enhancing the vision model's semantic comprehension and action execution capabilities. Further advancing the field, Brohan~\etal~\cite{brohan2023rt} and Li~\etal~\cite{li2023vision} have innovatively applied large language models to craft manipulation policies specifically for robotic applications.
This paper delves into the domain of embodied agents, focusing on the quadruped robot as a case study. It explores the challenges and opportunities in enabling quadruped robots to autonomously navigate and perform a variety of tasks as directed by human instructions, contributing to the broader discourse on robotic autonomy and intelligence.

\noindent\textbf{Robot Dataset.}
The realm of robotic learning has been at the forefront of advancing open-source datasets tailored explicitly for robot learning~\cite{bousmalis2018using,jiang2011efficient,yu2016more, shilane2004princeton,wu20153d,tang2023saytap,karnan2022scand}.
Within the domain of quadrupedal robots, previous works on datasets primarily focused on algorithms and leg controllers.
Eckert~\textit{et al.}\cite{eckert2019benchmarking} introduced a grading system and an online dataset for the collection and distribution of agility scores. Caluwaerts~\textit{et al.}~\cite{caluwaerts2023barkour} sought to address the lack of a standardized and scalable environment and inductive metrics. Tang~\textit{et al.}~\cite{tang2023saytap} pioneered the generation of desired joint position control using LLM.
Nevertheless, common limitations persist among these quadruped robot datasets: the quantity is insufficient and the tasks are limited in diversity.
Hence, our data repository aims to complement these endeavors. We curate and process an extensive array of skills on various objects, across diverse scenes in both the real world and virtual environments. The dataset includes image data, robot action data, and point cloud data, providing quadruped robots with rich perceptual information to tackle increasingly intricate and demanding tasks.

\section{Method} 

This section provides a detailed exposition of our proposed methodology. Initially, we present the definition of our proposed \textbf{QUAR-VLA} in Section~\ref{sec:3.1}.
Following that, Section~\ref{sec:3.2} outlines the collecting process of \textbf{QUARD}.
Lastly, we delve into the overarching architecture of \textbf{QUART} in Section~\ref{sec:3.3}.

\subsection{Problem Setup}
\label{sec:3.1}

The objective of \textbf{QUAR-VLA} is to construct a vision-language-action model learned from large-scale demonstration data and generate actions for closed-loop robot control. 

An overview of our \textbf{QUAR-VLA} is shown in Fig.~\ref{fig: teaser}. 
Our goal is to train a conditional policy \textbf{QUART} that can interpret RGB image(s), denoted $s \in \mathcal{S}$, together with a task instruction $w \in \mathcal{W}$, which correspond to a language string.
The policy is a mapping from images and instructions to actions, and can be written as $\mu: \mathcal{S} \times \mathcal{W} \rightarrow \mathcal{A}$, where the action space $\mathcal{A}$ consists of the 11-dimensional high-level commands as well as a terminate signal.
\begin{align}
\left[v_x, v_y, \omega_z, \theta_1, \theta_2, \theta_3, f, h_z, \phi, s_y, h_z^f, t \right]
\end{align}
Here, $v_x$, $v_y$, and $\omega_z$ represent the velocities along the x-axis, y-axis, and z-axis respectively. $\theta_1$, $\theta_2$, and $\theta_3$ indicate the gait pattern, $f$ denotes the frequency, $h_z$ represents the height of the robot, $\phi$ denotes the pitch angle, $s_y$ corresponds to the foot width, $h_z^f$ represents the foot height, and $t$ indicates the termination signal of the action.

Notably, in this study, we employ the discretization method proposed by \cite{brohan2023rt1} to discretize all continuous dimensions into 256 uniformly sized bins. This choice of discretization is motivated by the desire to reduce the complexity of action search and improve the stability and convergence of algorithms.

\begin{table*}[t!]
 \vspace{-1.em}
\caption{
\textbf{Tasks Definition. }
The "Type" means different capabilities of robots.
The "Level" devides the difficulty into 3 levels.
The "Skill" means different skill/task categories. 
The "Episode" signifies the number of experiments conducted for each task, which also corresponds to the number of trajectories.  
The "Description" is the illustration of the tasks.}

\label{tab:task_definition}
\vspace{-0.6em}
\renewcommand\arraystretch{1.4}
\tiny 
\centering
\setlength{\tabcolsep}{0.55mm}{
\begin{tabular}{p{0.12\columnwidth}p{0.07\columnwidth}p{0.2\columnwidth}p{0.08\columnwidth}p{0.48\columnwidth}} 
\hline
\textbf{Type}  & \textbf{Level} & \textbf {Skill}  & \textbf{Episode} & \textbf{Description} \\
\hline
Perception & Easy & Distinguish \textit{Letter} (sim) & 10K & Identify the correct one from multiple boxes with different printed letters \\
\hline
Basic   & Medium & Go to \textit{Object} (sim) & 72K & Navigate to the object and stop in front of it\\
Navigation  & Medium & Go to Object (real)  & 3K & Navigate to the object and stop in front of it \\
  \hline
 & Hard & Go through \textit{Tunnel} (sim)  & 48K & \textbf{Spatial
Navigation}: Go through the correct tunnel from two tunnels with different colors and shapes \\
 Advanced \qquad Capability & Hard & Go to \textit{Object} and avoid the obstacle (sim)  & 63K & \textbf{Obstacle
Avoidance}: Navigate to the object without colliding with the obstacle\\
 & Hard & Crawl under \textit{Bar} (sim)  & 1K &
\textbf{Environment Adaptation}: Crawl under a bar with a low height\\
 
  & Hard & Unload \textit{Object} into \textit{Receptacle} (sim)  & 52K & \textbf{Object
Manipulation}: Move with a ball on the back and unload it into a receptacle\\

\hline
 & & Total  & 259K & The total number of episodes\\

\hline

\end{tabular}
}
\vspace{-1.em}
\end{table*}
\vspace{-1.em}
\begin{figure}[t]
\begin{center}
  \includegraphics[width=0.8\columnwidth]{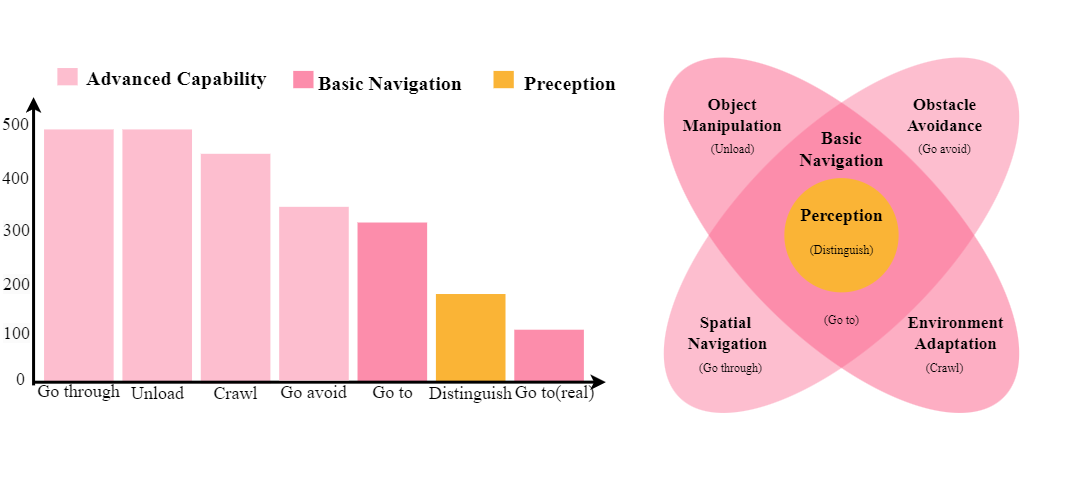} 
\end{center}
\vspace{-2.em}
\caption{
The left figure illustrates the trajectory lengths corresponding to different tasks and the right figure illustrates the relationships between tasks. As the difficulty of the skill increases, it can be observed that the average trajectory length gradually increases. The right figure demonstrates the relationship between the types of tasks: perception as foundational capability; basic navigation ability is built on perception; object manipulation, obstacle avoidance, spatial navigation, and environment adaption extend from perception and navigation. 
} 
\vspace{-1.em}
\label{fig:task_definition}
\end{figure}

\subsection{Large-scale Quadruped Robot Datasets}
\label{sec:3.2}
To enable an imitation learning system to generalize to new tasks with zero demonstrations of said task, we must be able to easily collect a diverse dataset, provide corrective feedback, and evaluate many tasks at scale. Therefore, we collect a large-scale multi-task dataset, \textbf{QUA}druped \textbf{R}obot \textbf{D}ataset (\textbf{QUARD}), which includes multiple tasks such as perception, basic navigation, and advanced capability like object manipulation.

\begin{figure}[t]
\begin{center}
  \includegraphics[width=1\columnwidth]{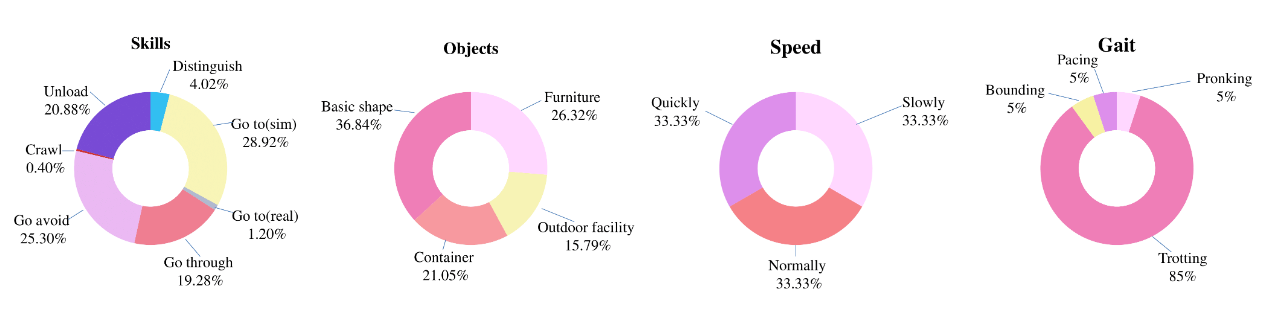} 
\end{center}
 \vspace{-1.5em}
\caption{Statistic analysis of \textbf{QUARD}. 1) Simulation data accounts for a larger proportion compared to real data. 2) Trotting occupied most of the share of gait. 3) Most tasks own similar episodes while Distinguish Letter and Go to Object (in reality) owns much less. 4) 3 types of speed occupied 1/3 separately.
}
\vspace{-1.em}
\label{fig:statistical}
\end{figure}

\noindent\textbf{Task Definition.} 
As is shown in Table~\ref{tab:task_definition}, there are seven tasks distributed across three levels of difficulty: easy, medium, and hard. The easy and medium tasks are designed such that the robot can accomplish them using fundamental skills. For instance, the ``Go to'' task is a basic action integral to all navigation tasks, while the ``Distinguish'' task serves as the foundational action for all perception tasks.
Hard tasks necessitate the robot to execute tasks using more advanced skills. A case in point is the ``Unload object'' task, which involves the robot distinguishing the target container, navigating to the target position, adjusting body position, and considering potential collisions to successfully unload the object. This process integrates perception, navigation, and whole-body manipulation. The complexity of tasks is reflected in the average trajectory length, with more challenging tasks requiring a greater number of steps and consequently taking longer to complete. The specific trajectory lengths and task distribution for each task are detailed in Fig.~\ref{fig:statistical} and Fig.~\ref{fig:task_definition}.

\noindent\textbf{System Setup.}
The robot used to collect the trajectory data is WR-2, which is a quadruped robot with 12 joints. WR-2 has around 25cm in standing height and 40cm body length. The command output is sent to the low-level command tracking controller (pre-trained command-conditioned policy in~\cite{pmlr-v205-margolis23a}) to generate the actual joint action of the robot. The simulation data was collected in Nvidia's Isaac Gym~\cite{makoviychuk2021isaac}, a powerful simulator that allows us to collect massive robot trajectories in parallel. 
The real data was collected by manually manipulating a quadruped robot in the lab setting. The perception data is provided by a RealSense d435 camera installed in the front of the robot.

\noindent\textbf{Data Collection.}
As to simulation data, A* and D* algorithms are used to plot optimal paths for a dog navigating through various objects and obstacles. The A* algorithm seeks the most cost-effective path, while the D* algorithm adapts to changes in real time. A Proportional-Derivative controller then converts these paths into target velocities for the dog, ensuring smooth movement. The combination of both algorithms provides a flexible and efficient method for path planning in various environments. For real data, The real data set is obtained in a laboratory environment using remote control.

\noindent\textbf{Consistency constraints.}
To maintain consistency between simulation and reality, we've established constraints for data collection. The robot starts at the origin, with the target randomly positioned within [2.7,3.3] meters on the x-axis and [0.9,1.1] meters on the y-axis. In obstacle scenarios, obstacles are placed 1.5 meters from the target's x-coordinate, with the same y-coordinate. Task termination criteria vary. For "go to object", "go to the object and avoid obstacle", and "crawl under bar" tasks, success is when the robot is less than one meter from the target. Other task success criteria are:
1) ``Unload object'': successful when the object is in the target container.
2) ``Go through tunnel'' and ``Crawl under bar'': successful when the robot reaches a specific position behind the tunnel or bar.
3) ``Distinguish letter'': successful when the robot correctly orients towards the visual target.

\noindent\textbf{More Statistics.}
The data collection setup adheres to a predefined language template as depicted in Fig.~\ref{fig:statistical}, where the task, target object, speed, and gait are explicitly defined.
To enhance diversity and generalization capabilities, the dataset includes a variety of common indoor furniture and outdoor facilities, in addition to basic shapes. The basic shapes come in four color variations: green, red, blue, and yellow. Examples of indoor furniture include bookshelves, ovens, and vases, while outdoor facilities encompass trashcans and benches.
Recognizing the distinct nature of motion robots as compared to manipulation robots, we also require robots to perform the same task with varying gaits and speeds to adapt to different instructions and environments. The corresponding ratios of gaits and speeds are presented in  Fig.~\ref{fig:statistical}.
More details about the datasets can be found in the supplementary material.

\begin{figure}[t]
\begin{center}
    \includegraphics[width=0.95\columnwidth]{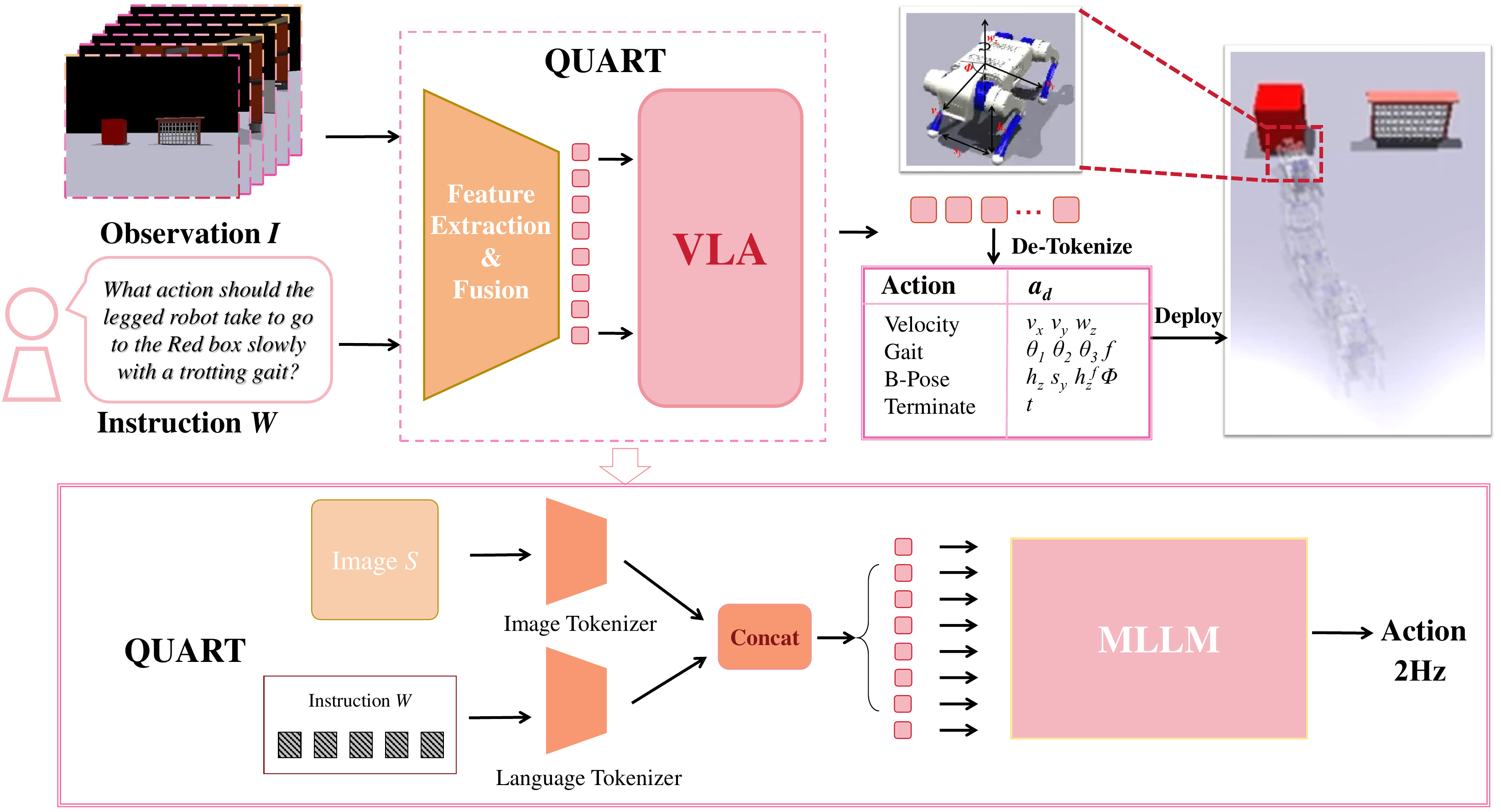}
\end{center}
\vspace{-2.em}
\caption{\textbf{Architecture of QUART}. It is designed to leverage the scene comprehension capability of a pretrained MLLM. It receives visual information as observation, and outputs an action representing the actual action taken by the robot based on text-form instructions, and de-tokenizes it into specific action values. \textbf{QUART} can generate a complete action sequence at a processing rate of 2Hz in actual scenarios, and hand it over to the underlying low-level strategy for execution.}
\vspace{-1.5em}
\label{fig:quart}
\end{figure}
\subsection{Vision-Language-Action Model }
\label{sec:3.3}

In comparison to VLA models~\cite{shah2023vint,stone2023openworld, brohan2023rt1, brohan2023rt2, gu2023rttrajectory} in other field, our research stands out due to the abundance of kinematics information available in this domain. This allows the robot to not only excel at goal-oriented navigation tasks but also perform a diverse range of whole-body manipulation operations with enhanced flexibility in gaits and body control. Next, we will present our \textbf{QUA}druped \textbf{R}obotic \textbf{T}ransformers \textbf{(QUART)}.

\textbf{QUART} relies on a pre-trained vision-language-model ~\cite{BavishiFuyu} to associate tokens from the model's existing tokenization with the discrete action bins. It is worth noting that training MLLMs to override existing tokens with action tokens is a form of symbol tuning~\cite{wei2023symbol}, which has been shown to work well in prior work. For ~\cite{BavishiFuyu}, integers up to 1000 each have a unique token, so we simply associate the action bins with the token representing the corresponding integer.
Notably, \textbf{QUART} model takes a single image $s$ and a natural language instruction $w$ as input, which are first converted into corresponding tokens~$t$ through a tokenizer~$\tau(t|s,w)$ and fed into a decoder-only transformer module to obtain discretized action tokens~$p(a_d|t)$. 

The policy \textbf{QUART} could be shown as follow:

\begin{equation}
\begin{aligned}
 &\operatorname{QUART}(a_d|s, w) = p(a_d|t) \tau(t|s, w)\\
\end{aligned}
\end{equation}

where $w,s$ are the input images and language instruction and $\tau$ represents the tokenizer and $p$ indicates the vision-language model to output action $a_d$.

\noindent\textbf{Action Detokenize.}
To directly convert models' output to valid robot actions for downstream control, we need detokenize the discrete action token $a_d$ into continuous representation $a_c$ (except for the discrete termination command). 

\begin{equation}
\begin{aligned}
 &a_c = \operatorname{Detokenize}(a_d)
\end{aligned}
\end{equation}

\noindent\textbf{Loss.}
We use a standard categorical cross-entropy objective and causal masking that was utilized in prior Transformer-based controllers~\cite{reed2022generalist,lee2022multi}.


\noindent\textbf{Inference Speed.}
In contrast to many applications of large models, such as natural language or image generation, one of the unique requirements for a model that needs to run on real robots in real-time is fast and consistent inference speed. 
It is important to highlight that our research specifically emphasizes the command control of quadruped robots. In contrast to low-level motor control, the command control outputs alleviate the strict control frequency requirements, enabling smooth integration of larger models and unlocking improved reasoning capabilities.
For
\textbf{QUART}, the inference time could get 2Hz.
More details about the deployment could be found in the supplementary material.


\section{Experiments} \label{sec:method}


We concentrate our experiments on the multi-task ability, generalization, and sim2real scaling.
and try to address the following questions:
1. How effective is the vision-language-action architecture for multi-task quadruped task compared to previous VLM baselines?
2. How well do these models generalize to unseen semantic attributes like object shape, color, and unseen verbal information?
3. How do these simulation data contribute to real-scene performance?
More details about the model architecture and training strategy can be found in the supplementary material.

\subsection{Implementation Details}

\noindent\textbf{Training Details.}
We train a specific instantiation of \textbf{QUART}, derived from 8B pre-trained VLM models~\cite{BavishiFuyu} for superior performance. 
And we use learning rate {2e-5} and batch size 256 to fine-tune the model for 100K gradient steps. 
Both models are trained with the next token prediction objective, which corresponds to the behavior cloning loss in robot learning.

\noindent\textbf{Evaluation Details.}
We follow the standard robot evaluation metrics~\cite{brohan2023rt2,brohan2023rt1}, success rate (SR), to evaluate the overall performance.
The signal we used for robot control is the 11-dimensional command information with the action space outlined in Section~\ref{sec:3.1} and 1 stop terminal command means the end of action. More details for each experiment can be seen in the supplementary material.

\noindent\textbf{Baselines.}
Considering the absence of VLA models work on quadruped robots at present, we have taken the following baselines into account for a fair comparison: \textbf{CLIP}\cite{radford2021learning}, \textbf{R3M}\cite{nair2022r3m}, and \textbf{VC-1}\cite{vc2023}. 

1. \textbf{CLIP} \cite{radford2021learning}uniquely encodes both textual and visual data, fusing these embeddings to create feature sets that integrate textual and visual information.

2. \textbf{R3M} \cite{nair2022r3m}is a visual representation model derived from the Ego4D dataset, which acts as a unified, static perception module for policy learning applications.

3. \textbf{VC-1} \cite{vc2023}pushes the boundaries of visual representation learning by amalgamating data from multiple datasets and assessing performance across a diverse range of tasks via the implementation of CortexBench.

Since the aforementioned models do not include language conditioning, we include this aspect by separately embedding the language command, allowing us to compare it to our method. Specifically, we concatenate the resulting language embedding tokens with the image tokens generated by the vision model and pass the combined token sequences through a policy head to produce action outputs.

\subsection{Overall Performance}

\begin{figure}[t]
\begin{center}
      \includegraphics[width=1\columnwidth]{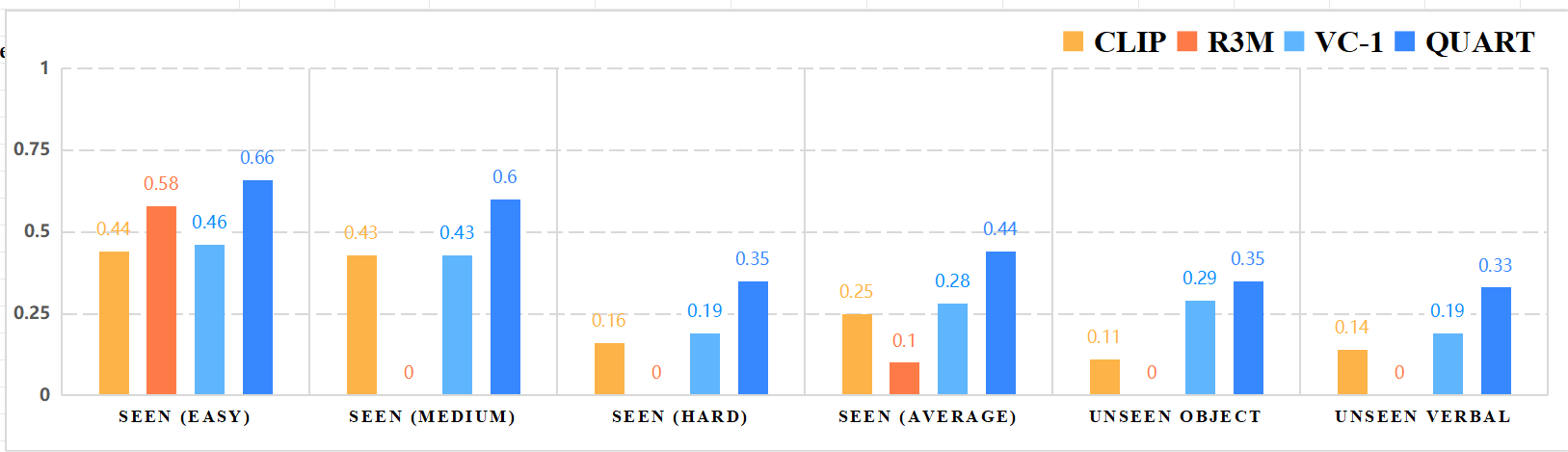} 
\end{center}
\vspace{-1.2em}
\caption{\textbf{Tasks Successful Rate. }{R3M lacks alignment with text, which means that although it has some recognition capabilities, it struggles to understand actions through instructions. CLIP and the VC-1 possess the ability to align text and images, enabling them to generate and execute actions based on instructions and observations.As shown above, while R3M may have a slight advantage in easy seen tasks (distinguish), it performs poorly in more complex tasks requiring action execution, significantly lagging behind CLIP and VC-1. \textbf{QUART} fully leverages the advantages of large models, enabling semantic understanding of images and alignment with textual information.} }
\label{fig:unseen_dpx}
\end{figure}

\begin{table*}[t!]
\caption{\textbf{Overall performance.} \textbf{QUART} has achieved success rates far exceeding those of the baselines in tasks of all difficulty levels, especially in the most challenging crawl and unload tasks, where the baselines have no record of success.}
\label{tab:exp_baseline}
\renewcommand\arraystretch{1.2}
\scriptsize
\centering
\begin{tabular}{cc|c|c|c|c|c|c|c|c }
\hline
\multicolumn{2}{c|}{\thead{}} 
&\multicolumn{6}{c|}{\thead{Seen}} 
&\multicolumn{2}{c}{\thead{Unseen}}
\\
\cline{3-10}
\multicolumn{2}{c|}{\thead{}} 
&\multicolumn{1}{c|}{\thead{Easy}} 
&\multicolumn{1}{c|}{\thead{Medium}}
& \multicolumn{4}{c|}{\thead{Hard}}
&\multicolumn{1}{c|}{\thead{}}
&\multicolumn{1}{c}{\thead{}}
\\
\multicolumn{2}{c|}{\thead{}} 
&\multicolumn{1}{c|}{\thead{Distinguish}} 
&\multicolumn{1}{c|}{\thead{Go to}}
& \multicolumn{1}{c}{\thead{Go avoid}}
&\multicolumn{1}{c}{\thead{Go through} }
&\multicolumn{1}{c}{\thead{Crawl}} 
&\multicolumn{1}{c|}{\thead{Unload}}
&\multicolumn{1}{c|}{\thead{Object}}
&\multicolumn{1}{c}{\thead{Verbal}}
\\
\hline
{{CLIP}\cite{radford2021learning}}&
&0.44 & 0.43 & 0.45& 0.19& 0 & 0& 0.11 &0.14\\

{R3M}\cite{nair2022r3m}&
&0.58 &0 &0 &0 & 0 &0 &0 &0\\
{VC-1}\cite{vc2023}&
&0.46 &0.43 &0.45 &0.31 &0  &0 &0.29 & 0.19\\


\textbf{QUART}&
&\textbf{0.66}  &\textbf{0.60}& \textbf{0.53}& \textbf{0.41}& \textbf{0.32}& \textbf{0.12} & \textbf{0.35} &\textbf{0.33}\\

\hline

\end{tabular}
\vspace{-1.em}
\end{table*}

\noindent\textbf{Multi-task Performance.}
To assess the overall capabilities of VLA models on multiple tasks, we conduct evaluations using instructions randomly selected from the training set. It's important to note that this evaluation introduces variations in the placement of objects and other setup factors (such as robot position). This variation demands the system's ability to generalize effectively to realistic environmental variability. In total, over 1500 episodes are tested in this evaluation, comprising 425 episodes for going to objects, 500 for going to objects without colliding with the obstacle. 150 for going through the tunnel, 100 for unloading objects, 100 for distinguishing objects, and 75 for crawling under the bar.

\textbf{1. Comparison within VLM baselines.}
The experiment results reveal that R3M has poor performances on tasks except distinguish. The primary reason appears to be the lack of alignment with language semantics, which hinders its ability to comprehend tasks beyond basic discrimination. In contrast, CLIP and VC-1 demonstrate satisfactory performance on fundamental perceptual tasks, such as navigation (e.g., "go to" commands). However, their efficacy diminishes significantly when tasks require complex mechanical movements. This observation suggests that while visual language models (VLMs) can grasp abstract principles of the world, directly applying VLMs does not readily translate to the execution of intricate physical tasks.

\textbf{2. VLM baselines vs QUART.}
As is shown in Table~\ref{tab:exp_baseline}, Our model has achieved optimal performance on nearly all the baseline models. In our study, we have achieved a decoder-only VLA framework. This approach diverges from the traditional single-layer MLP policy head by leveraging the sequential nature of action generation. This allows for the implicit learning of interdependencies between different action dimensions through the use of a transformer. Consequently, while the performance gains may be marginal in simple tasks, there is a noticeable enhancement in tasks that involve complex mechanical movements.
significantly improved its perceptual capabilities by incorporating commonsense from the multi-modal large model (MLLM).

\begin{figure*}[t]
\begin{center}
\vspace{0.0em}
  \includegraphics[width=0.8\textwidth]{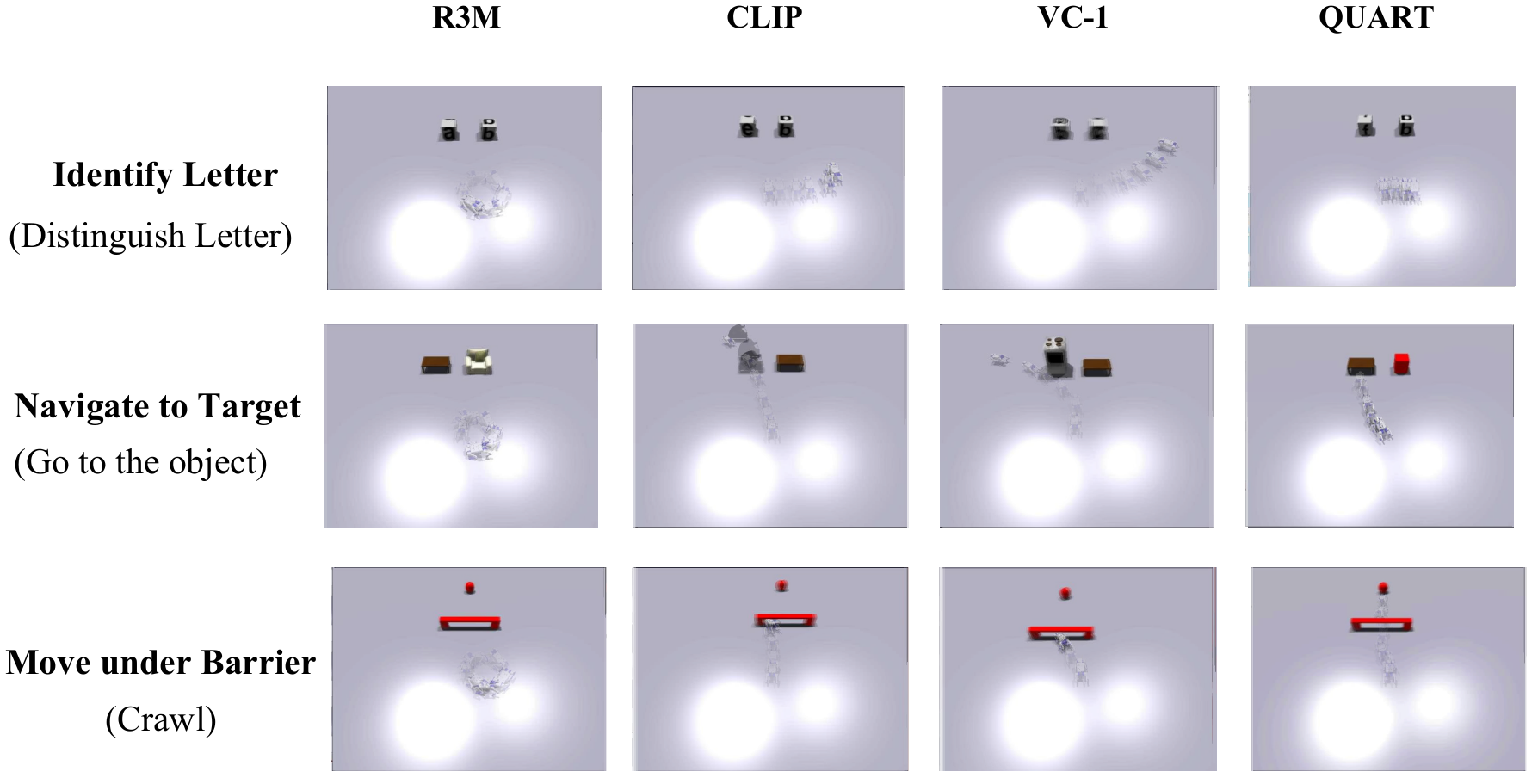} 
\end{center}
\vspace{-1.8em}
\caption{\textbf{ Cases with unseen verbal.}
Here are 3 verbal instructions. 
The words below represent the names of instructions in the training data, with the words above in bold indicating verbal instruction.
When confronted with unseen instructions, the alighment between the existing language and the integration of vision and action cues within the baselines is compromised, resulting in task failure. 
This failure manifests in behaviors such as repetitive motion, misdirection, wrong terminate commands. 
Conversely, \textbf{QUART}, leveraging the language prowess inherited from large language models, adeptly achieves generalization under novel instructions, thereby effectuating the harmonization of vision, language, and action.} 

\label{fig:seen}
\vspace{-1.5em}
\end{figure*}
\begin{figure}[t]
\begin{center}
      \includegraphics[width=1\columnwidth]{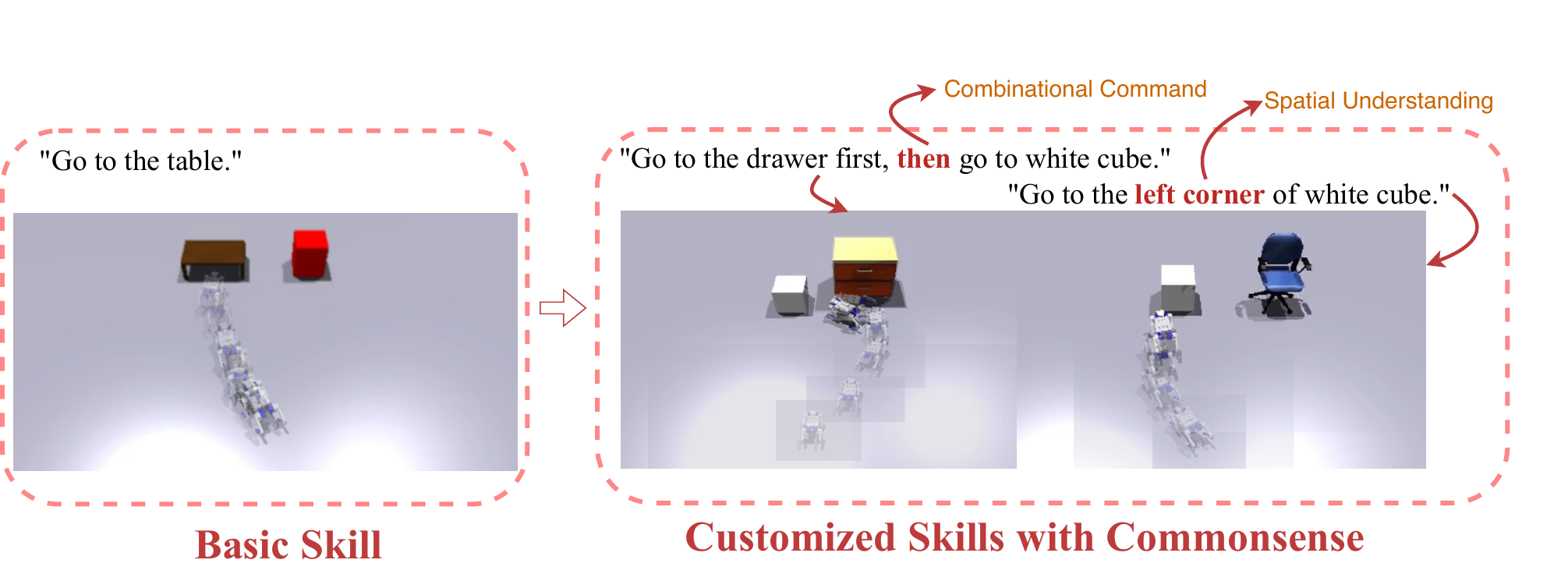} 
\end{center}
\vspace{-1.5 em}
\caption{
\textbf{QUART} can extend basic skills to accommodate complex customized skills with commonsense. It shows example performances of combinational command and spatial understanding towards customized command. 
}
\vspace{-1.5 em}
\label{fig:visulization}
\end{figure}

\begin{table*}[t!]
\caption{\textbf{Scaling experiments on real-world execution of QUART. }\textbf{QUART} was trained on three training schemes that mix simulated and real data. The experimental results show that as the amount of simulated data increases, the success rate of \textbf{QUART} also rises, demonstrating well scaling capabilities.}
\label{tab:exp_sim2real}
\vspace{-1em}
\renewcommand\arraystretch{1.1}
\scriptsize
\centering
\setlength{\tabcolsep}{0.42mm}{
\begin{tabular}{c|c|c|c}
\hline
\multicolumn{1}{c|}{\thead{Sim Data : Real Data}} 
&\multicolumn{1}{c|}{\thead{0K :3K}}
&\multicolumn{1}{c|}{\thead{25.6K: 3K}}
&\multicolumn{1}{c}{\thead{256K: 3K}}
\\
\hline
\textbf{QUART}
& 3/20 & 7/20 &13/20\\
\hline

\end{tabular}
}
\vspace{-1.em}
\end{table*}

\begin{figure}[t]
\begin{center}
  \includegraphics[width=1\columnwidth]{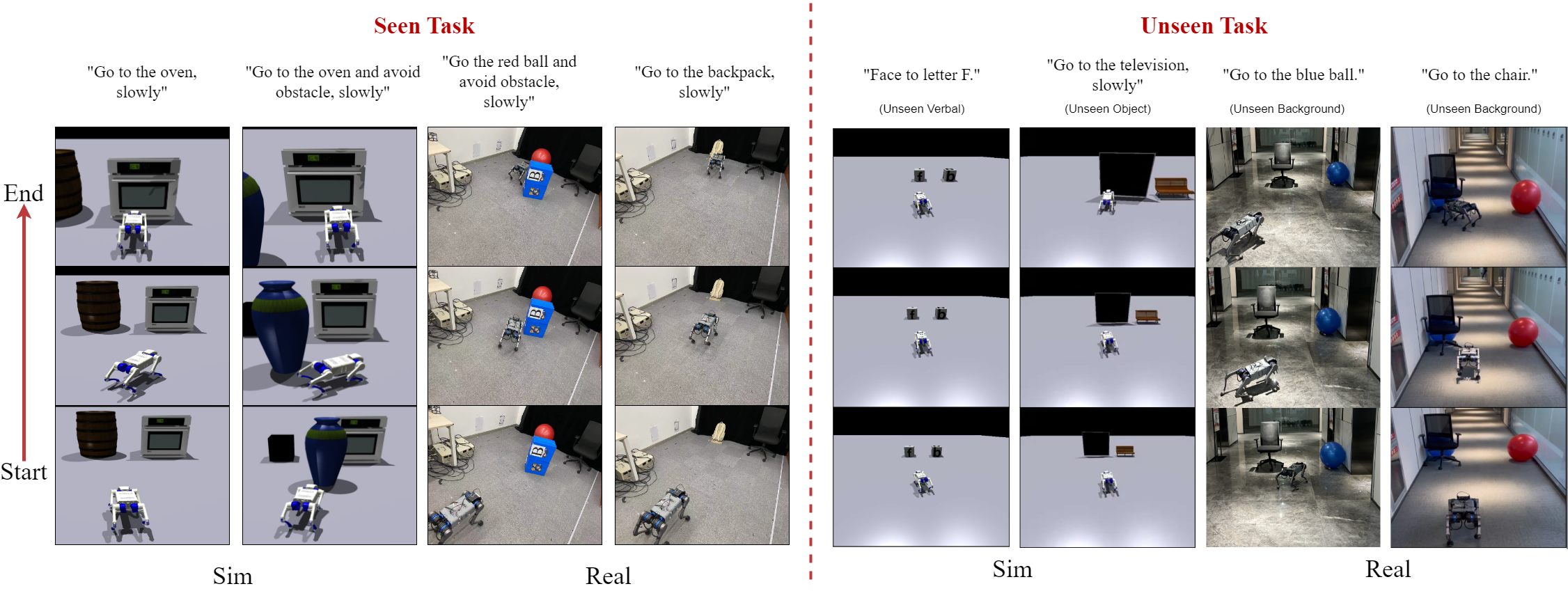} 
\end{center}
 \vspace{-1.5em}
\caption{
\textbf{ Visualization of seen and unseen tasks.} This figure demonstrates the excellent performance of \textbf{QUART} for seen tasks in both simulated and real-world environments. The left part shows 4 instances of seen tasks in both simulation environment and real-world. The right part shows performances on unseen tasks, which contain unseen verbal, object, and background.}
\label{fig:unseen}
 \vspace{-1.5em}
\end{figure}

\noindent\textbf{Generalization capabilities.}
To examine the adaptability in unanticipated scenarios, we orchestrated two primary examinations: one concentrated on unfamiliar objects, and the other on unprecedented linguistic directives. For the unfamiliar objects, we cast our gaze upon an array of circumstances: objects belonging to an identical category but exhibiting divergent textures and colors; objects from the same category but of disparate shapes; and objects with differences in shapes and textures, absent from current datasets.
Turning towards the unprecedented linguistic instructions, we characterized unfamiliar instances as linguistic directives that bear the same semantics but vary in expression. For example, within the "\textit{go to the object}" task, we gauged the directive's adaptability by employing "\textit{navigate to target}" as the testing instruction.


\textbf{1. Unseen Object.}
As depicted in Fig.~\ref{fig:unseen_dpx}, we can observe the enhanced generalization ability of our \textbf{QUART} model.
For unseen objects, both \textbf{QUART} and VC-1 perform well, thanks to the advantages of the pre-trained VLM model. However, when it comes to unseen verbal descriptions, apart from \textbf{QUART}, all other models essentially failed.  This is attributed to the inherent advantage provided by the pre-trained Multimodal Large Language Model (MLLM), enabling the recognition of distinct variations in object types and the comprehension of subtle semantic nuances. Consequently, the model exhibits exceptional generalization performance.

\textbf{2. Unseen Verbal instruction.}
The training directives utilized were relatively monotonous, demonstrating that minor alterations in instruction can often perplex the model, thereby revealing an imperative for generalization across language directives.
Initially, the difficulties introduced by merely replacing descriptions of skill for task generalization were investigated. Evident from Table~\ref{tab:exp_baseline} and Fig.~\ref{fig:seen} is that \textbf{QUART} surpasses other baseline methods, an innate advantage of large-scale multimodal models.
In addition, to explore more intriguing linguistic directives and fully harness the universal capabilities of the Large Language Model (LLM), Fig.~\ref{fig:visulization} presents instances where the model demonstrates a clear comprehension and application of combined directives constructed via "first" and "then". It was further noted that even in the absence of explicit directional information within the dataset, the model reliably interpreted linguistic directives which, in turn, influenced task execution accuracy.
The ability to utilize language for precise control of quadruped robots facilitated by scene perception, as showcased in Fig.~\ref{fig:visulization}, is of immense significance. More intriguing examples and analyses can be found in the supplementary materials.

\noindent\textbf{Sim2Real Transferring.}
Due to the scarcity of authentic data, a considerable amount of simulated data has emerged as a reliable resource for training machine learning models. This has been particularly beneficial in the field of robotics, where real-world data can be challenging and costly to collect. Simulated data, generated from physics engines and computer graphics, can provide diverse and scalable environments for robots to learn from. In this context, we evaluate the performance of joint training using both simulated and real data in real-world scenarios. This approach, often referred to as sim-to-real transfer, leverages the strengths of both types of data. Simulated data provides a wide range of scenarios for initial training, while real-world data ensures the model's applicability to authentic environments.
As illustrated in Table~\ref{tab:exp_sim2real}, our experiments show that the performance of the \textbf{QUART} model improves consistently with the incorporation of simulated data. This demonstrates the effectiveness of simulated data in enhancing the model's ability to handle real-world tasks. The joint training approach, which combines broad simulation data with narrow real-world data, enables \textbf{QUART} to work effectively in real-world environments. The broad simulation data provides a comprehensive training ground, while the narrow real-world data ensures the model's robustness and adaptability to real-world scenarios. For more detailed experimental results and further insights, please refer to the supplementary materials provided.

\section{Conclusion \& Future Work} \label{sec:conclusion}

This paper emphasizes the significance of deploying Vision-Language-Action models on quadruped robots. 
We introduce the concept of \textbf{V}ision-\textbf{L}anguage-\textbf{A}ction tasks for \textbf{QUA}druped \textbf{R}obots \textbf{(QUAR-VLA)}, which seamlessly integrates visual information and instructions from diverse modalities to generate executable actions.
Specifically, \textbf{QUAR-VLA} focuses on two main aspects of deploying VLA models for quadruped robots: defining the action space to balance flexibility and efficiency, and addressing the scarcity of large-scale training datasets. To tackle these two questions, we present the \textbf{QUART} models tailored for quadruped robots and the \textbf{QUARD} dataset, which includes diverse tasks such as navigation and manipulation.
Our extensive evaluation shows that our approach leads to performant robotic policies and enables \textbf{QUART} to obtain a range of emergent capabilities. This includes generalization to novel objects, the ability to interpret commands not present in the robot training data, and the ability to perform rudimentary reasoning in response to user commands.

Furthermore, this work has these areas requiring further enhancements: \\
\textbf{1) Automated Data Collection for Larger Datasets.} Future work can explore automated processes for data collection, involving techniques such as active learning and data augmentation can be employed to further expand the dataset size and diversity.
\textbf{2) Better solution for sim2Real gap.}  Integrating domain adaptation methods like domain randomization, data augmentation, and more accurate modeling of physics engines will be a better solution compared with direct co-training.
\textbf{3) Improving Inference Speed for Real-time Control.} Future works will explore hardware acceleration techniques and model compression techniques to enable faster and more efficient execution of the models. 

\title{Supplementary material:\\QUAR-VLA: Vision-Language-Action Model for Quadruped Robots}

\author{Pengxian Ding$^{12}$ \quad Han Zhao$^{1}$ \quad Wenxuan Song$^{1}$ 
\quad Wenjie Zhang$^{1}$ \\ \quad Ningxi Yang$^{1}$ \quad Donglin Wang$^{1}$\thanks{Corresponding author}}
\institute{$^{1}$Westlake University \quad $^{2}$Zhejiang University \\ 
\email{\{dingpengxiang,zhaohan34,wangdonglin\}@westlake.edu.cn}} 

\maketitle




%

%
\section{Abstract}

In this supplementary material, we will provide the following supplementary information:

\textbf{Extended Data Collection Details}: We will present additional details regarding the data collection process, including information about the specific environments used and the criteria for determining successful data acquisition for different tasks.

\textbf{Enhanced Experimental Results}: We will provide more comprehensive information about the experimental setup, including a comparative analysis of the results obtained in both seen and unseen environments. Furthermore, we will compare the performance of different model architectures to provide a more thorough evaluation.

\textbf{Expanded Deployment Results}: To facilitate real-world deployment, we will present additional experimental results showcasing the model's performance across a broader range of tasks and scenarios.

\textbf{Extended Visualizations}: We will include supplementary visual results that highlight various failure cases observed in both real-world and simulation settings, offering a more comprehensive understanding of the model's limitations and areas for improvement.

\section{Details of data collection}
\textbf{Environment}: We prioritize the richness of tasks. In terms of environmental setup, our current tasks are all carried out in relatively simple scenarios without complex visual background information for data collection and model evaluation. At the same time, all experiments are conducted on flat terrains, and we have not yet conducted experiments on complex terrains. Our next goal is to use simulation environments with higher visual fidelity, and add some backgrounds that are more in line with real-world scenarios and more varied terrains for data collection.


\textbf{Planning Algorithm}: 
In simulation scenarios, we adopted a large-scale parallel simulation environment for the need of rapid automated data collection. During data collection, we used traditional path search algorithms to implement the robot's route selection. In the early stage of data collection, we adopted the D* algorithm. After evaluating the quality of the early data, we found that the paths planned by D* tend to generate curves with larger turning angles. This could potentially lead to the robot frequently losing sight of the target object, thus affecting the quality of the strategy obtained by imitation learning. Therefore, we changed the navigation algorithm to A*. Taking the go\_avoid task as an example, we directly access the positions of the robot, the target object, and obstacles through the simulator. The initial centroid coordinates of the robot on the x-y plane are used as the starting point of the path, and the centroid coordinates of the target object are used as the endpoint. We then add the areas of obstacles on the map. After the path planning is completed, a PD controller is used to convert the path into current velocity and direction. Since we set the speed of the path collection to three levels (fast, normal, slow), the obtained speed needs to be processed to roughly satisfy the speed range set by the levels.

\textbf{Excution Details}: 
In control systems, the combination of high-frequency and low-frequency control is a strategic approach to achieve finer control and optimize system performance. This integration can be described by a simple mathematical relationship: $N=f_{high}/f_{low}$, where $f_{high}$ represents the high-frequency control rate, $f_low$ represents the low-frequency control rate, and N is the ratio between the two.
In this equation, 
N indicates the number of times the high-frequency control needs to be executed within a given low-frequency period. 
The advantage of this combination is that it allows the system to achieve rapid responses and finer adjustments while maintaining the stability and efficiency of the low-frequency control. High-frequency control is typically used to address the rapid changes in system states, while low-frequency control focuses more on long-term system stability and energy management.

\section{More experiments}

\subsection{Detailed results on seen tasks}
Firstly, we supplement two additional frameworks: \textbf{VLA(LLaVa)} (which uses LLaVa as the backbone for multi-modal large models) and 
\textbf{Aligned VLM+P} \textbf{(Transformer)} (policy head is a decoder-only transformer) to validate the rationality of our VLA model framework design;
Next, we provide the results trained on a single task to verify the performance advantages on multi-task;
Then, we present more granular results for each task.

\noindent\textbf{Two additional frameworks}:


\textbf{1. Experiment Setting:}

1) \textbf{VLM baseline}: In the main paper, we reproduced previous baselines based on VLM by utilizing features extracted from VLM in conjunction with a policy head to generate actions. However, the policy head at that time was merely in the form of an MLP, called \textbf{VLM+P(MLP)}, and we did not explore the differences between various types of policy heads, particularly the decoder-only transformer (\textbf{VLM+P(Transformer)}).Therefore, we have additionally supplemented this study with different policy-head layers to ascertain whether the limitations of the VLM+Policy approach are related to the design of the policy head.

2) \textbf{VLA architecture}: Furthermore, in the paper, we experimented with a multi-modal large model based on the fuyu-8b model(\textbf{VLA(Fuyu)}) to verify whether the effectiveness of VLA is related to the choice of multi-modal large model foundations. To this end, we have also presented experiments based on the LLaVa(\textbf{VLA(LLaVa)}) to validate the benefits of a VLA approach grounded in multi-modal large models.
\begin{table*}[t!]
\caption{\textbf{Detailed performance on different architecture.}}
\label{tab:different_architecture}
\renewcommand\arraystretch{1.2}
\scriptsize
\centering
\begin{tabular}{c|c|c|c|c|c|c|c }
\hline
\multicolumn{1}{c|}{\thead{}} 
&\multicolumn{1}{c|}{\thead{}} 
&\multicolumn{1}{c|}{\thead{Distinguish}} 
&\multicolumn{1}{c|}{\thead{Go to}}
& \multicolumn{1}{c}{\thead{Go avoid}}
&\multicolumn{1}{c}{\thead{Go through} }
&\multicolumn{1}{c}{\thead{Crawl}} 
&\multicolumn{1}{c}{\thead{Unload}}
\\
\hline
{Unaligned VLM+P(MLP)}&{{CLIP}\cite{radford2021learning}}&
0.44 & 0.43 & 0.45& 0.19& 0 & 0\\

Aligned VLM+P(MLP)&{VC-1}\cite{vc2023}&
0.46 &0.43 &0.45 &0.31 &0  &0 \\
{Aligned VLM+P(Transformer)}&{RT-1}\cite{vc2023}&
0.22 &0.15 &0.4 &- &0  &0 \\
\hline
{VLA(Fuyu)}&\textbf{QUART}&
\textbf{0.66}  &\textbf{0.60}& \textbf{0.53}& \textbf{0.41}& \textbf{0.32}& \textbf{0.12} \\

{VLA(LLaVa)}&\textbf{QUART}&
\textbf{0.66}  &\textbf{0.52}& \textbf{0.40}& \textbf{0.32}& \textbf{0.18}& \textbf{0.16} \\
\hline

\end{tabular}
\vspace{-1.em}
\end{table*}



\textbf{2. Experiment Analysis:}

\textbf{1) Unaligned VLM + Policy(MLP)}: For the R3M model, the visual backbone employed is a ResNet-50, which has not been aligned with textual features. Consequently, this model is only capable of performing the most rudimentary tasks within its scope.

\textbf{2) Aligned VLM + Policy(MLP)}: In the case of CLIP and VC-1, the visual and textual features have been aligned, enabling the models to comprehend and execute simple tasks. They perform reasonably well on tasks such as "go to," "go through," and "go avoid," which do not involve manipulation of the robot's body. The primary reason for this adequate performance is that these tasks only require changes in velocity along the x-axis and yaw orientat.

\textbf{3) Aligned VLM + Policy(Transformer)}:
As is shown in Table~\ref{tab:different_architecture}, we have also referred to the RT-1 paradigm, employing different policy heads (\textbf{VLM} \textbf{+P}\textbf{(Transformer)}) to ascertain whether the limitations are inherent to the MLP architecture. We can observe that even when the policy is switched to a Decoder-only Transformer, the trend of the RT-1 method remains consistent with the previous VLM+P (MLP) approach. Only tasks that involve simple distinction and those related to the velocity of the aircraft's center of mass have success rates; tasks such as crawl and unload are still not achievable. This demonstrates that the choice of different policy heads does not affect the performance of the VLM+policy paradigm.

\textbf{4)Vision + Language + Action (VLA)}: Within the VLA framework, we have utilized the entire decoder-only VLM backbone. However, we directly map action instructions to the language space. During inference, each dimension of our action (e.g., leg width) engages in joint reasoning with previously inferred information (e.g., robot height). This approach allows for the implicit learning of associations between different action dimensions, thereby effectively grasping the coordinated relationships between multiple parts of the robot and performing well on more complex tasks (e.g., crawling).

To investigate whether different multi-modal large models (MMLMs) affect our model's performance, we introduce two variants: Fuyu-8B(used in main paper) and LLaVa-7B. The primary distinction between the two lies in the fact that the former encodes the original image directly, while the latter employs the widely used visual CLIP feature extraction module. From the results, we can see that there is not much difference in performance corresponding to which base model is used. This indicates the importance of the VLA paradigm.


\noindent\textbf{Multi-task vs Single-task Performance:}
To validate the performance of our multi-task learning approach, we conducted separate training on each individual task to ascertain the benefits that multi-task learning confers on the interrelated tasks.
As is shown in Table~\ref{tab:multi-task vs single-task}, It has been observed that, for both single and multi-task scenarios, the performance of multi-task training has yielded superior results in all but the simplest tasks(distinguish). This indicates that the paradigm of joint training in multi-task settings has enabled the learning of commonalities between different tasks, thereby underscoring the necessity of multi-task co-training.

\noindent\textbf{Detailed Performance:}
As is shown in Table~\ref{tab:detailed_result}, we present detailed results for the seen tasks.

\begin{table*}[t!]
\caption{\textbf{Multi-task performance vs Single-task performance.}}
\label{tab:multi-task vs single-task}
\renewcommand\arraystretch{1.2}
\scriptsize
\centering
\begin{tabular}{c|c|c|c|c|c|c }
\hline
\multicolumn{1}{c|}{\thead{}} 
&\multicolumn{1}{c|}{\thead{Distinguish}} 
&\multicolumn{1}{c|}{\thead{Go to}}
& \multicolumn{1}{c|}{\thead{Go avoid}}
&\multicolumn{1}{c|}{\thead{Go through} }
&\multicolumn{1}{c|}{\thead{Crawl}} 
&\multicolumn{1}{c}{\thead{Unload}}
\\
\hline
{CLIP-Multi}\cite{radford2021learning}&
0.44 & \textbf{0.43} & \textbf{0.45}& \textbf{0.19}& 0 & 0\\
{{CLIP-Single}\cite{radford2021learning}}&
\textbf{0.52} & 0.34 & 0.37& 0.04& 0 & 0\\

\hline
{VC-1-Multi}\cite{vc2023}&
0.46 &\textbf{0.43} &\textbf{0.45} &0.31 &0  &0 \\
{VC-1-Single}\cite{vc2023}&
\textbf{0.70} &0.37 &0.40 &\textbf{0.34} &0  &0 \\
\hline
\textbf{QUART-Multi}&
\textbf{0.66 } &\textbf{0.60}& \textbf{0.53}&\textbf{ 0.41}&\textbf{ 0.32}& \textbf{0.12} \\
\textbf{QUART-Single}&
0.30  &0.36& 0.19& 0.30& 0.25& 0.08 \\


\hline

\end{tabular}
\vspace{-1.em}
\end{table*}

\begin{table*}[t!]
\caption{\textbf{Detailed results on seen tasks.}}
\label{tab:detailed_result}
\renewcommand\arraystretch{1.2}
\scriptsize
\centering
\begin{tabular}{cc|c|c|c }
\hline
\multicolumn{2}{c|}{\thead{}} 
&\multicolumn{1}{c|}{\thead{Distinguish letter c}} 
&\multicolumn{1}{c|}{\thead{Distinguish letter d}}
& \multicolumn{1}{c}{\thead{Go to cooker}}
\\
\hline
{{CLIP}\cite{radford2021learning}}&
&0.36 & 0.52 & 0.36\\

{VC-1}\cite{vc2023}&
&0.48 &0.44 &0.36 \\

\textbf{QUART}&
&\textbf{0.76}  &\textbf{0.56}& \textbf{0.72}\\
\hline

\hline
\multicolumn{2}{c|}{\thead{}} 
&\multicolumn{1}{c|}{\thead{Go to ball}} 
&\multicolumn{1}{c|}{\thead{Go to cube}}
& \multicolumn{1}{c}{\thead{Go to oven}}

\\
\hline
{{CLIP}\cite{radford2021learning}}&
&0.56 & 0.24 & \textbf{0.56}\\

{VC-1}\cite{vc2023}&
&\textbf{0.64} &0.40 &0.32 \\

\textbf{QUART}&
&0.60  &\textbf{0.64}& 0.44\\

\hline
\multicolumn{2}{c|}{\thead{}} 
&\multicolumn{1}{c|}{\thead{Go avoid cooker}} 
&\multicolumn{1}{c|}{\thead{Go avoid drawers}}
& \multicolumn{1}{c}{\thead{Go avoid fan}}
\\
\hline
{{CLIP}\cite{radford2021learning}}&
&0.44 & \textbf{0.52} & 0.28\\


{VC-1}\cite{vc2023}&
&0.44 &\textbf{0.52} &0.20 \\

\textbf{QUART}&
&\textbf{0.44}  &0.48& \textbf{0.36}\\

\hline
\multicolumn{2}{c|}{\thead{}} 
&\multicolumn{1}{c|}{\thead{Go avoid sofa}} 
&\multicolumn{1}{c|}{\thead{Go through triangle tunnel}}
& \multicolumn{1}{c}{\thead{Go through rectangle tunnel}}

\\
\hline
{{CLIP}\cite{radford2021learning}}&
&0.56 & 0.04 & 0.24\\
{VC-1}\cite{vc2023}&
&0.64 &\textbf{0.40} &0.28 \\

\textbf{QUART}&
&\textbf{0.84}  &0.24& \textbf{0.47}\\

\hline

\hline
\multicolumn{2}{c|}{\thead{}} 
&\multicolumn{1}{c|}{\thead{Crawl gate}} 
&\multicolumn{1}{c|}{\thead{Unload traybox}}
& \multicolumn{1}{c}{\thead{Average}}

\\
\hline
{{CLIP}\cite{radford2021learning}}&
&0 & 0 & 0.25\\

{VC-1}\cite{vc2023}&
&0 &0 &0.28 \\

\textbf{QUART}&
&\textbf{0.32}  &\textbf{0.04}& \textbf{0.44}\\
\hline
\end{tabular}
\vspace{-1.em}
\end{table*}

\subsection{Detailed results on unseen tasks}

In the context of unseen tasks, we conducted experiments to assess the model's performance on novel objects and unseen language instructions. The novel objects are categorized into three types: objects of the same category but with different shapes; objects of the same shape but with different colors; and entirely different objects. The unseen language instructions involve paraphrasing existing descriptions with synonymous terms to test the robustness of the model's performance.

\noindent\textbf{Detailed Performance on unseen objects:}
In the experiments, we use the {yellow, red, green and blue} as four base color in the seen tasks, and use the {gold, pink, orange, purple} as the unseen color. For each objects appear in the seen tasks, we all test another object which is the same type but with different shape. We also test on objects which do not appear in the seen tasks: pillow, computer and window. As is shown in Table~\ref{tab:detailed_unseen_result}, we present detailed results for the unseen objects. We can see that our method have more generalization ability in unseen objects.

\begin{table*}[t!]
\caption{\textbf{Detailed results on unseen objects.}}
\label{tab:detailed_unseen_result}
\renewcommand\arraystretch{1.2}
\scriptsize
\centering
\begin{tabular}{c|c|c|c|c|c }
\hline

\multicolumn{1}{c|}{\thead{}} 
&\multicolumn{1}{c|}{\thead{Go to}}
& \multicolumn{1}{c}{\thead{Go avoid}}
&\multicolumn{1}{c}{\thead{Go through} }
&\multicolumn{1}{c}{\thead{Crawl}} 
&\multicolumn{1}{c}{\thead{Unload}}
\\
\hline
{{CLIP}\cite{radford2021learning}}&
 0.4 & 0.46& 0.19& 0.04 & 0\\

{VC-1}\cite{vc2023}&
0.38 &0.41 &0.36 &0  &0 \\

\hline

\textbf{QUART}&
 \textbf{0.4}& \textbf{0.73}& \textbf{0.41}& \textbf{0.35} & \textbf{0.01}\\
\hline

\end{tabular}
\vspace{-1.em}
\end{table*}

\begin{table*}[t!]
\caption{\textbf{Detailed results on unseen verbal information.}}
\label{tab:detailed_unseen_verbal_result}
\renewcommand\arraystretch{1.2}
\scriptsize
\centering
\begin{tabular}{c|c|c}
\hline

\multicolumn{1}{c|}{\thead{}} 
&\multicolumn{1}{c|}{\thead{Identify Letter}} 
&\multicolumn{1}{c}{\thead{Navigate to target}}
\\
\hline
{{CLIP}\cite{radford2021learning}}&
0.40 & 0.44 \\

{VC-1}\cite{vc2023}&
0.28 &0.48   \\

\hline

\textbf{QUART}&
0.40& \textbf{0.52}\\
\hline

\multicolumn{1}{c|}{\thead{}} 
&\multicolumn{1}{c|}{\thead{Move under barrier}} 
&\multicolumn{1}{c}{\thead{Deposit object into container}}
\\
\hline
{{CLIP}\cite{radford2021learning}}&
 0.12 & 0\\

{VC-1}\cite{vc2023}&
0 &0  \\

\hline

\textbf{QUART}&
\textbf{0.28} &\textbf{0.04}\\
\hline
\end{tabular}
\vspace{-1.em}
\end{table*}
\noindent\textbf{Detailed Performance on unseen verbal instruction:}
Here are alternative expressions for the tasks while maintaining the same meaning. 
Within the parentheses are the instructions for the seen tasks, followed by the modified instructions.
1. (Distinguish Letter) Identify Letter
2. (Go to the object) Navigate to target
3. (Crawl under the barrier) Move under barrier
4. (Unload the object) Deposit Object into container
As is shown is Table~\ref{tab:detailed_unseen_verbal_result}, We can observe that with instructions that carry the same semantics but different expressions, the performance of QUART significantly surpasses that of the baseline.

\subsection{More results of customized skills}

In the manuscript, we demonstrate the capability of our model to generalize to customized skills that were not present in the training tasks, such as complex spatial perception and the ability to combine tasks. Herein, we will present additional case studies to illustrate this skill further.
\begin{figure}[t]
\begin{center}
  \includegraphics[width=0.3\columnwidth]{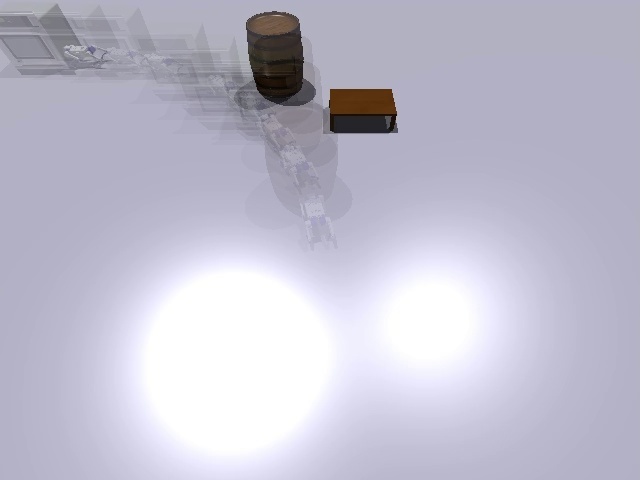} 
  \includegraphics[width=0.3\columnwidth]{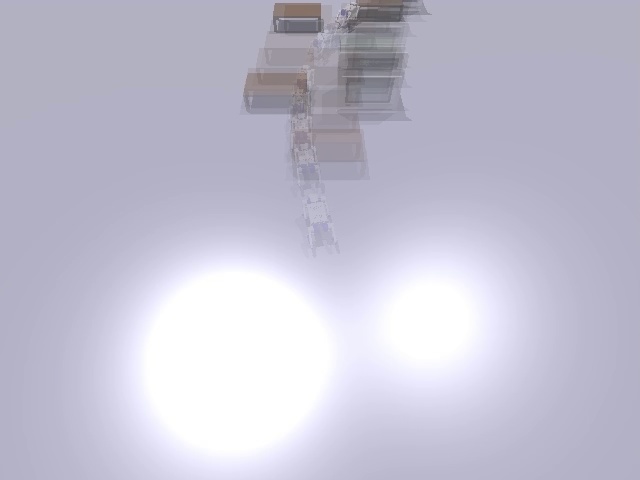}
  \includegraphics[width=0.3\columnwidth]{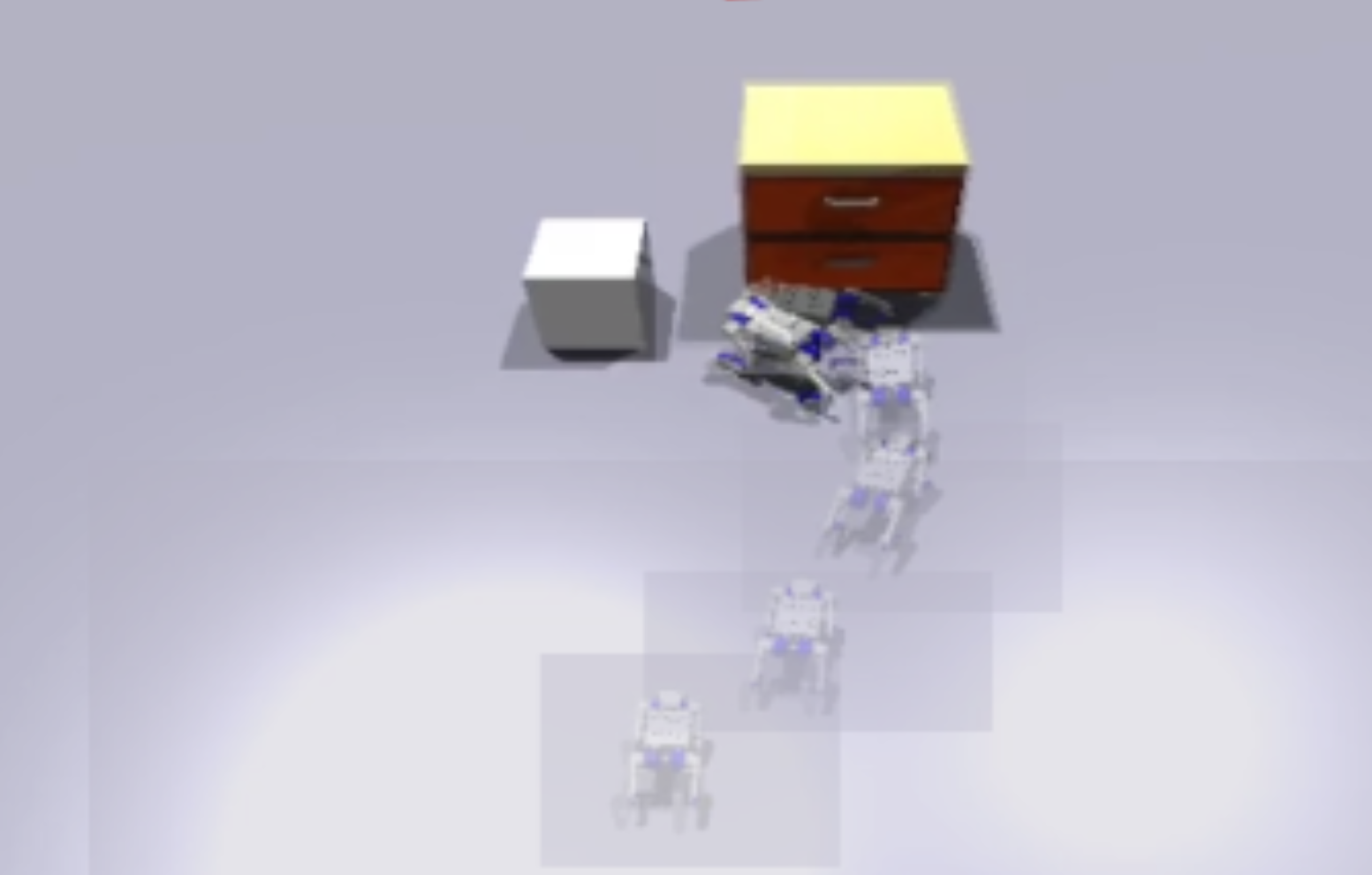}
\end{center}
\caption{Mission \textit{go to the left corner of the object}. The left picture is produced by model CLIP. The middle picture is produced by model VC-1. The right picture is produced by \textbf{QUART}.
}
\vspace{-1.em}
\label{fig:supp_corner}
\end{figure}

\begin{figure}[t]
\begin{center}
  \includegraphics[width=0.3\columnwidth]{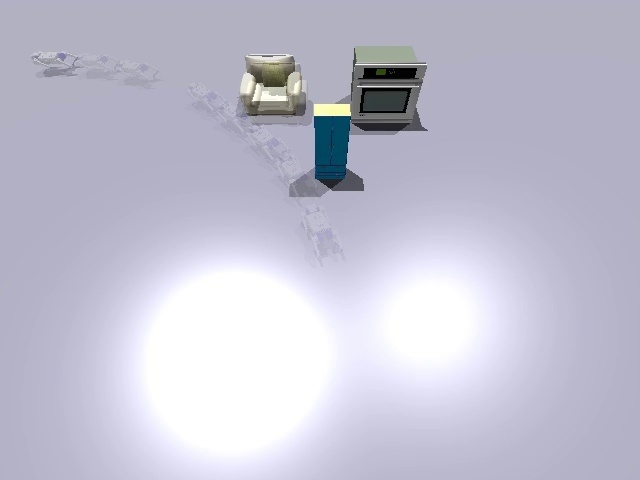} 
  \includegraphics[width=0.3\columnwidth]{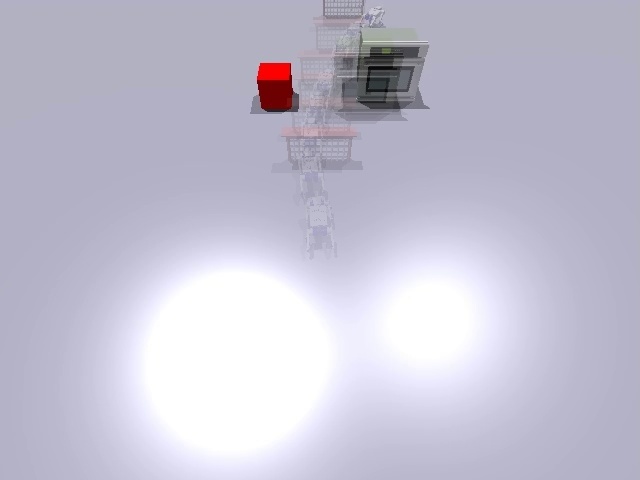}
  \includegraphics[width=0.3\columnwidth]{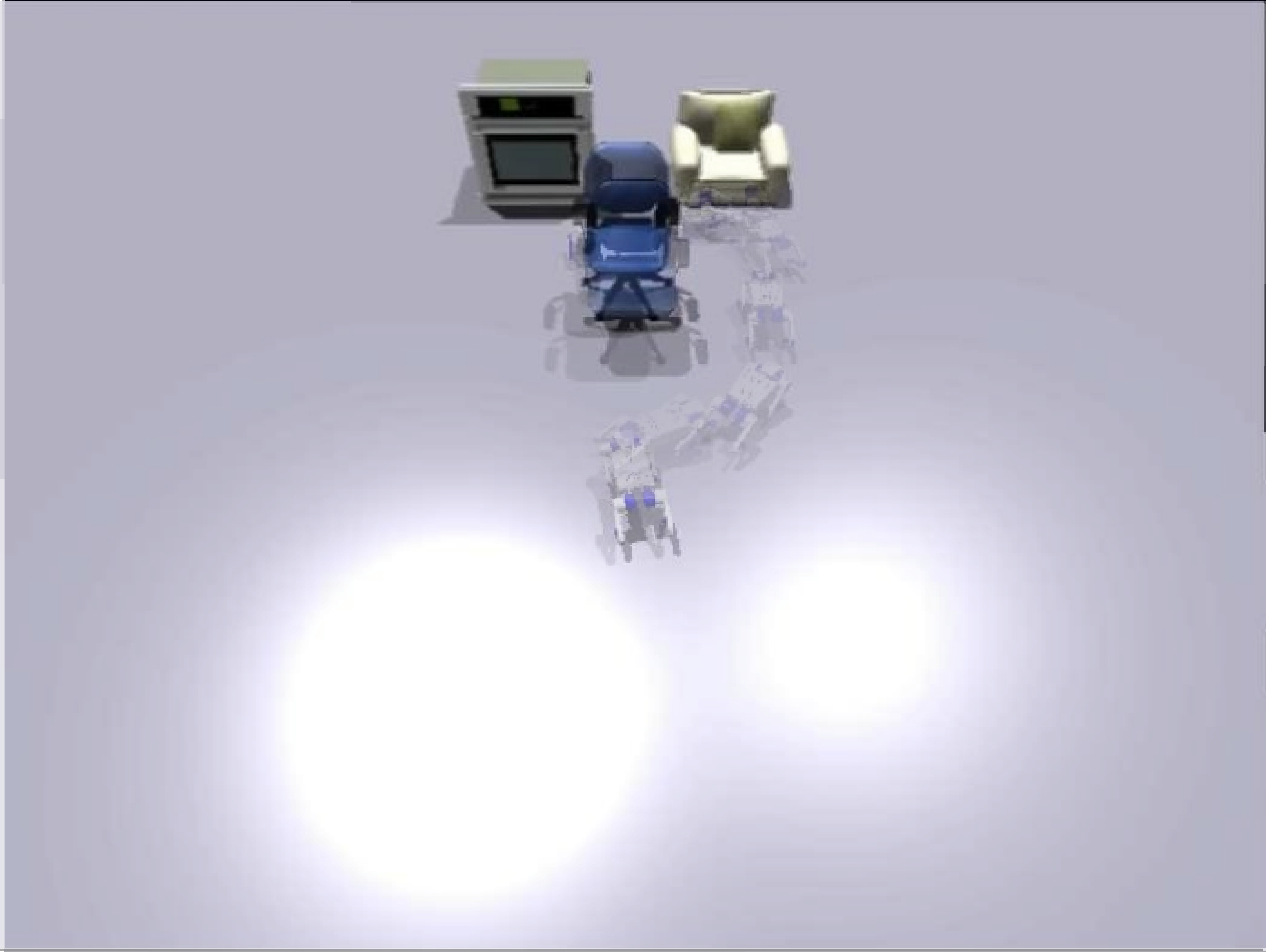}
\end{center}
 \vspace{-1.5em}
\caption{Mission \textit{go to the back of the object}. The left picture is produced by model CLIP. The middle picture is produced by model VC-1. The right picture is produced by \textbf{QUART}.
}
\vspace{-1.em}
\label{fig:supp_back}
\end{figure}

\begin{figure}[t]
\begin{center}
  \includegraphics[width=0.3\columnwidth]{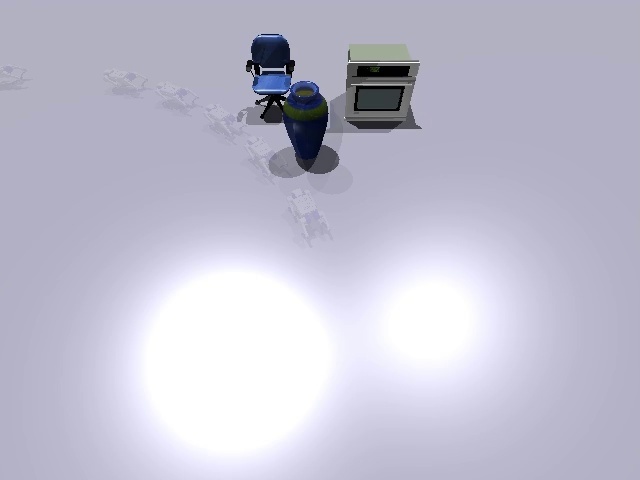} 
  \includegraphics[width=0.3\columnwidth]{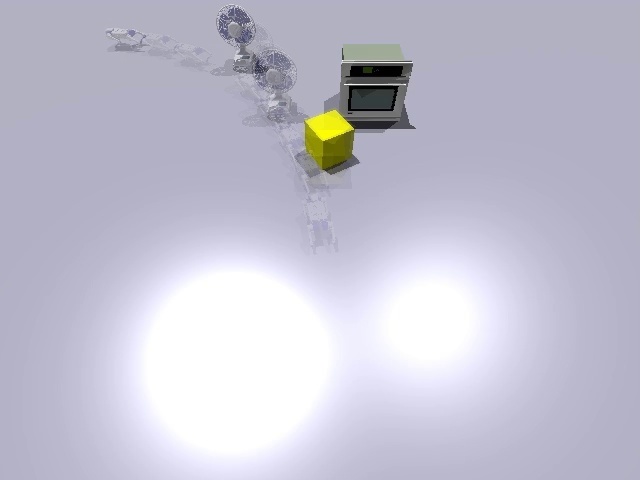}
  \includegraphics[width=0.3\columnwidth]{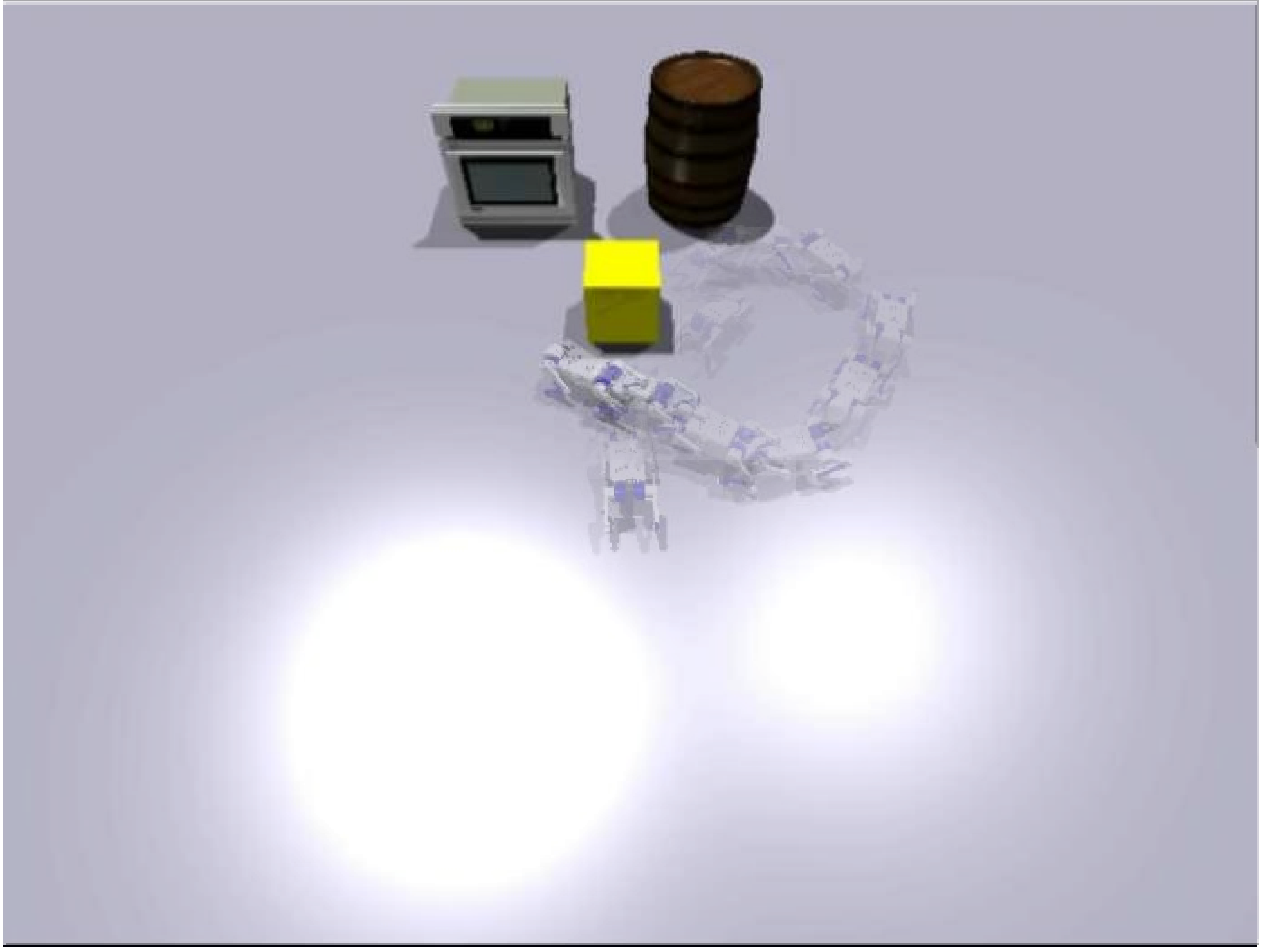}
\end{center}
\caption{Mission \textit{go to the right and the left}. The left picture is produced by model CLIP. The middle picture is produced by model VC-1. The right picture is produced by \textbf{QUART}.
}
\vspace{-1.em}
\label{fig:supp_rl}
\end{figure}

\textbf{Case1: Go the left corner of the object. Figure~\ref{fig:supp_corner}}

\textbf{Case2: Go the the back of object. Figure~\ref{fig:supp_back}}

\textbf{Case3: Go the the left and the to the right. Figure~\ref{fig:supp_rl}}

In these cases, we could find our model could understand the spatial relationships and have the ability to excute combinational skills.
\section{More analysis on real robot excution}

In the main paper, we previously conducted 20 trials on the "Go to" task and compiled the results, which indicated an enhancement in sim2real performance. In this section, we present further experimental outcomes of the model as the size of the simulation dataset increases, as well as results across a broader range of tasks, in order to substantiate the efficacy of our methodology. 

We show the real robot experiments from the following 5 aspects:
1. Effectiveness in seen scenes
2. Sim2Real transfer capabilities
3. Rubustness in different localization
4. Rubustness in different workspace
5. Rubustness in unseen scenes

More results can be found in https://sites.google.com/view/quar-vla/quar-vla-eccv24.



\section{Limitation and future work}
From the perspective of dataset composition, the current magnitude of data and the diversity of trajectories are insufficient. A potential direction for future work is to leverage GPT-based automation to generate more varied and enriched datasets. Additionally, the existing datasets lack complex terrains and long-horizon tasks; enhancing the complexity of the dataset is a crucial method for advancing quadrupedal tasks.

In terms of task formulation, the current input modalities are limited to visual and textual components. Exploring how to utilize additional modalities (e.g., LiDAR point clouds) to address issues that visual information alone cannot resolve, such as occlusion problems, is a direction worthy of investigation.
Furthermore, the existing models are not yet capable of more flexible control in terms of frequency. Although the high-level command action frequency can complete some tasks, more challenging tasks, such as pole crossing, require higher frequency control to achieve higher success rates. Therefore, accelerating the base model's speed and designing a reasonable sampling mechanism for high-frequency output is an essential component.

Of course, addressing the sim2real gap is key to effectively utilizing real-world data. The co-training approach adopted in this paper is based on the premise that the sim2real gap for large models is not significant. However, how to more efficiently employ various sim2real methods, such as domain adaptation and randomization, to solve the domain gap between the real and simulated domains is also a line of thought worth exploring. Lastly, given the substantial amount of sub-optimal data present in the data collection process, how to utilize this data and enable the large model to learn valuable knowledge from failures through reinforcement learning is an important future direction.

In summary, this is the inaugural work in extending multi-modal large models to quadrupedal robots. In response to the existing challenges of quadrupedal robots, we have designed a dataset that combines extensive simulated data with a small amount of real data for quadrupedal robot VLA and developed a framework based on large models to implement this task. This work has a certain catalytic effect on the development of the robotics community, and we hope for more suggestions to further refine this work in the future, thereby advancing the progress of mobile robotics.





%
%
\bibliographystyle{splncs04}
\bibliography{main}
\end{document}